\documentclass[letterpaper]{article} 
\usepackage{aaai2026}  
\usepackage{times}  
\usepackage{helvet}  
\usepackage{courier}  
\usepackage[hyphens]{url}  
\usepackage{graphicx} 
\usepackage{natbib}  
\usepackage{caption} 
\usepackage{amsfonts} 
\usepackage{algorithm}
\usepackage{algorithmic}

\usepackage{newfloat}
\usepackage{listings}
\usepackage{multirow}
\usepackage{subcaption}
\usepackage{booktabs}     
\usepackage{amsmath}      
\DeclareCaptionStyle{ruled}{labelfont=normalfont,labelsep=colon,strut=off} 
\floatstyle{ruled}
\newfloat{listing}{tb}{lst}{}
\floatname{listing}{Listing}
\nocopyright 

\title{Time2Agri: Temporal Pretext Tasks for Agricultural Monitoring}
\author{
  Moti Rattan Gupta \textsuperscript{\rm 1}
  Anupam Sobti \textsuperscript{\rm 1}
}
\affiliations{
    \textsuperscript{\rm 1}Plaksha University\\

    Mohali\\
    Punjab, India\\
    moti.gupta@plaksha.edu.in, anupam.sobti@plaksha.edu.in
}

\usepackage{bibentry}

\begin{document}

\maketitle

\begin{abstract}
  Self Supervised Learning(SSL) has emerged as a prominent paradigm for label-efficient learning,
  and has been widely utilized by remote sensing foundation models(RSFMs).
  Recent RSFMs including SatMAE, DoFA, primarily rely on masked autoencoding(MAE), contrastive learning or some combination of them. However, these pretext tasks often overlook the unique temporal characteristics of agricultural landscape, namely nature's cycle.
  Motivated by this gap, we propose three novel agriculture-specific pretext tasks, namely Time-Difference Prediction(TD), Temporal Frequency Prediction(FP),
  and Future-Frame Prediction(FF). Comprehensive evaluation on SICKLE dataset shows FF achieves 69.6\% IoU on crop mapping and FP reduces yield prediction error to 30.7\% MAPE, outperforming all baselines, and TD remains competitive on most tasks. Further, we also scale FF to the national scale of India, achieving 54.2\% IoU outperforming all baselines on field boundary delineation on FTW India dataset.
\end{abstract}


\begin{figure}[t]
    \centering
    \includegraphics[width=0.95\columnwidth]{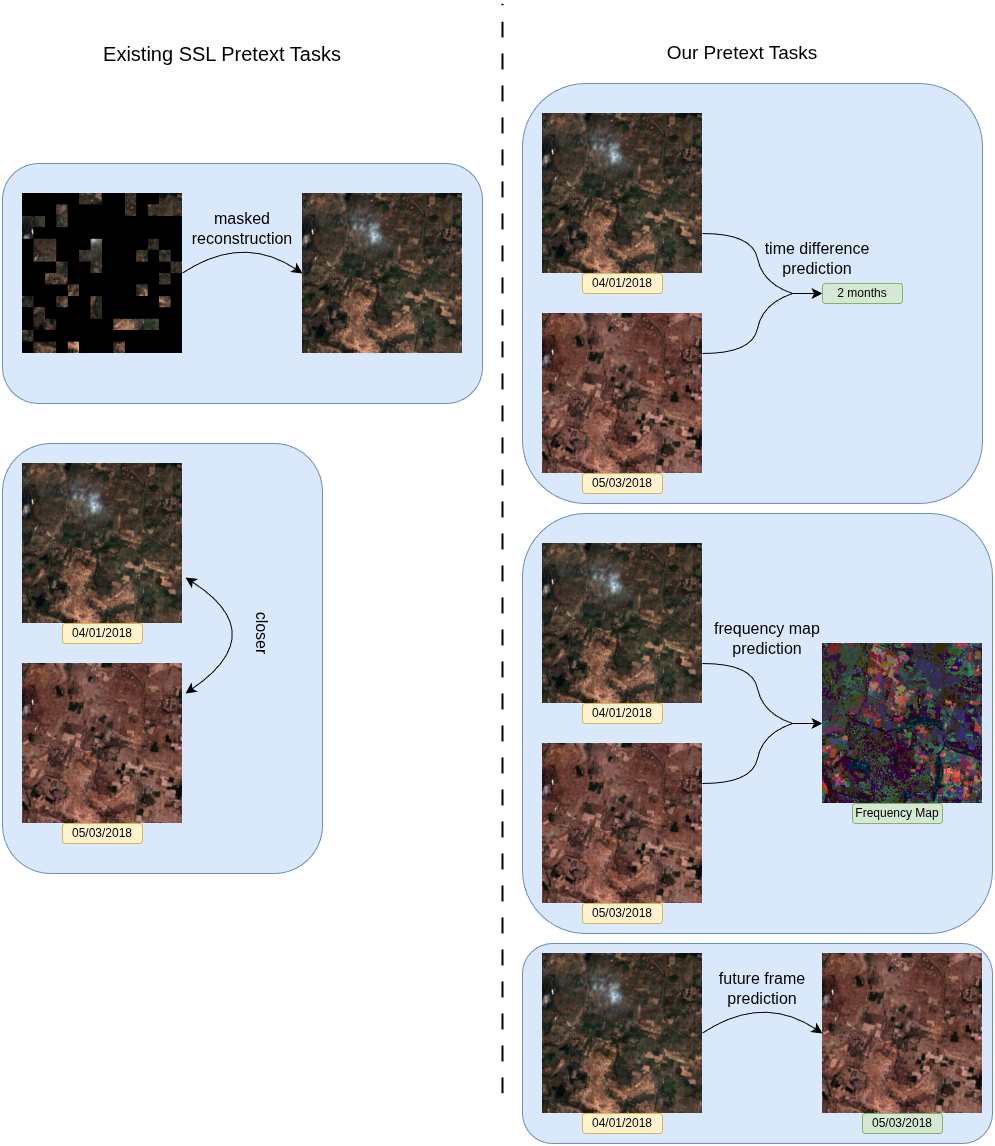}
    \caption{Overview of the proposed pretext tasks for self-supervised learning in agricultural remote sensing. On the right side, we have traditional pretext tasks focusing on masked reconstruction and invariance to augmentation(in case temporal), while on the right we have our temporally inspired pretext tasks.}
    \label{fig:pretext_overview}
\end{figure}

\begin{figure*}[ht]
  \centering
  \includegraphics[width=\linewidth]{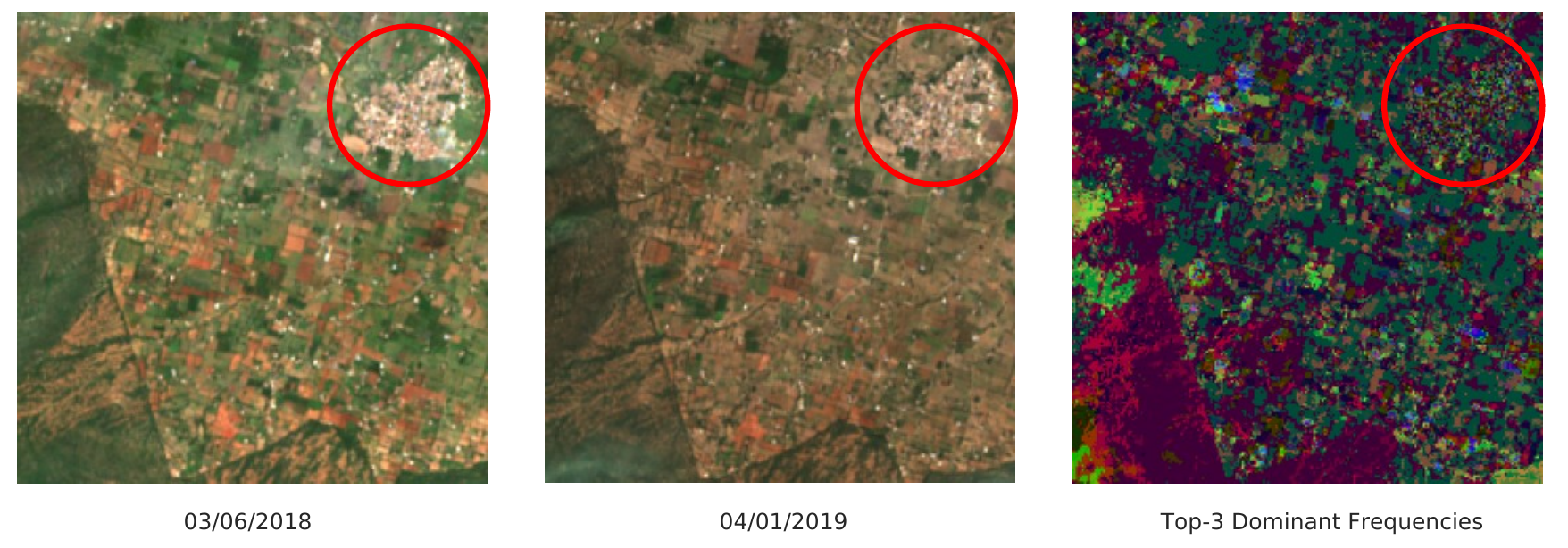}
  \caption{Comparison of a region across two different acquisitions. Agricultural areas exhibit large, coherent frequency clusters—represented by different colors—that correspond to varying crop growth rates and naturally delineate farm parcels. In contrast, urban regions(highlighted in red) do not display such structured clustering, indicating distinct temporal dynamics.}
  \label{fig:sample_analysis}
\end{figure*}

\section{Introduction}
\label{sec:intro}
With over 582 million people projected to be chronically undernourished by 2030, of which 53\% would be concentrated in Africa, the world is significantly off-track in achieving SDG 2 - Zero Hunger \cite{fao2024state}. These are further exacerbated by climate extremes like droughts and heatwaves affecting crop yields, and crop failures for farmers \cite{JRC139423}. Remote sensing plays a critical role in addressing the growing challenges of food security and agricultural sustainability, particularly within the framework of Agriculture 4.0, by enabling improved crop monitoring, yield forecasting, and resource optimization \cite{RSDSSAPACAP}. However, the lack of labelled data for training machine learning models, particularly in developing countries, remains a significant bottleneck in leveraging the full potential of remote sensing data for agriculture.

Remote Sensing Foundation Models (RSFMs), built upon the principle of self-supervised learning, have shown remarkable promise in learning from the abundant unlabelled satellite imagery data, demonstrating superior performance on downstream tasks including landcover mapping, crop type mapping, and field boundary delineation \cite{lu2025visionfoundationmodelsremote}. However, we argue most of them are suboptimal for agricultural monitoring due to the following reasons:
\begin{enumerate}
    \item \textit{Lack of domain-specific knowledge}: RSFMs trained primarily using masked autoencoding and/or contrastive learning do not incorporate domain-specific knowledge for agricultural monitoring.
    \item \textit{Lack of regional focus}: RSFMs trained on global datasets may not be optimal for regional agricultural monitoring due to differences in crop types, farming practices, and climate.
\end{enumerate}

Inspired by work of \citet{pmlr-v235-rolf24a}, who argue that satellite imagery constitutes a distinct modality within computer vision due to its unique spatial, spectral, and temporal properties, we extend this perspective to agricultural monitoring. Specifically, we argue that agricultural landscapes exhibit characteristic temporal dynamics driven by the natural cycle of sowing, growth, and harvest. These seasonal cycles strongly shape the spectral and temporal signatures, leading to predictable and recurring patterns which exhibit strong spatial clustering. In contrast, urban landscapes typically exhibit weaker spatial clustering(see Figure~\ref{fig:sample_analysis}). We argue that this inherent regularity in agricultural data provides a strong inductive bias that can inform the design of more effective pretext tasks for self-supervised learning, ultimately enabling the development of richer and more task-relevant representations for agricultural monitoring.

To this end, we propose three novel pretext tasks that leverage the seasonal nature of agricultural land and explain our intuitions behind them,
\begin{enumerate}
    \item \textit{Temporal Difference Prediction}: This task predicts the time difference between two images from a bitemporal pair of satellite images. The intuition is that agricultural landscapes undergo predictable transformations over time—from bare soil to seedlings to mature crops to harvest—and the magnitude of visual change correlates strongly with the elapsed time. By learning to estimate temporal gaps, the network implicitly learns to recognize and differentiate between various phenological stages, enabling it to capture the progression of crop development cycles. 
    \item \textit{Temporal Frequency Prediction}: This task predicts the per-region dominant temporal frequencies from a bitemporal satellite image pair. Agricultural landscapes exhibit characteristic periodic patterns—annual cycles for most crops, biannual patterns for double-cropping systems, and multi-year rotations. By learning to predict these frequency characteristics from just two observations, the network learns to identify crop types and farming practices based on their inherent temporal dynamics, capturing information that extends far beyond the immediate observation period. 
    \item \textit{Future Frame Prediction}: Given a bitemporal pair of satellite images, this task predicts the future image from the past image. This approach exploits the highly predictable nature of agricultural cycles—if we know the current crop stage and the time of year, we can reasonably predict what the field will look like in the future. Unlike urban landscapes, agricultural areas follow managed, cyclical patterns. This task forces the network to understand the causal relationships between current conditions and future states, learning rich representations of crop phenology and management practices.
\end{enumerate}

We evaluate our proposed pretext tasks across a range of key agricultural monitoring tasks at regional scale of Tamil Nadu, on the SICKLE benchmark \cite{Sani_2024_WACV}. Furthermore, we assess the performance of one of our best performing pretext task on the field boundary delineation at the national scale of India, on FTW India\cite{kerner2024fieldsworldmachinelearning}. We have observed consistent improvements over both state-of-the-art remote sensing foundation model DoFA \cite{xiong2024neuralplasticityinspiredmultimodalfoundation} and a self-supervised MAE baseline. Our results suggest that incorporating domain-specific cues in the pretext task design can result in rich representations for agricultural monitoring tasks.

Beyond pretext task design, we have also investigated the impact of geographical scale for training agricultural models. Specifically, we compare models trained at a national scale with those trained at a regional scale. Our findings reveal that regional scale pretraining outperform national scale pretraining on most regional downstream tasks, emphasizing the importance of regional agricultural representations.
To summarize, our contributions are as follows:
\begin{enumerate}
    \item We propose three novel pretext tasks for self-supervised learning that leverage the seasonal nature of agricultural land.
    \item We demonstrate the effectiveness of the proposed pretext tasks on crop type mapping, crop yield prediction, sowing, transplanting and harvesting date prediction at regional scale of Tamil Nadu, and on field boundary delineation task at national scale of India.
    \item We investigate the impact of geographical scale on pretraining for agricultural monitoring, and show that regional scale pretraining outperforms national scale pretraining for tackling regional agricultural challenges.
\end{enumerate}

To the best of our knowledge, this is the first work to leverage domain-specific agricultural knowledge in the design of pretext tasks for agricultural monitoring. We further demonstrate that region-specific pretraining outperforms national-scale approaches, underscoring the importance of both informed task design and geographic focus in advancing Agriculture 4.0.

\section{Related Work}

\label{sec:related}
\paragraph{Pretext Tasks in Satellite Imagery}
Early remote sensing SSL pretext tasks adapted their natural image counterparts, focusing on invariance of spatially closer tiles\cite{Jean2018Tile2VecUR} or patches\cite{9140372}, while others extended colorization \cite{vincenzi2020colorspacelearningselfsupervised}. Later, methods like GASSL\cite{Ayush_2021_ICCV} and SeCo\cite{Manas_2021_ICCV} utilized the temporal dimension of satellite imagery for invariance. CaCo \cite{Mall_2023_CVPR} built upon SeCo to ensure invariance only to seasonal changes. However, this invariance to seasonal change is suboptimal for agricultural landscapes, as seasonality is an important marker of agricultural land, and can lead to loss of rich phenological features.(see Section~\ref{sec:intro})

\paragraph{Masked Autoencoders in Remote Sensing}
Recent RSFMs have build upon the MAE\cite{He2021MaskedAA} framework, including SatMAE \cite{Cong2022SatMAEPT}, ScaleMAE \cite{reed2023scalemaescaleawaremaskedautoencoder}, SelectiveMAE \cite{wang2024scalingefficientmaskedimage} and M3AE \cite{li2024maskedangleawareautoencoderremote}, while some combine masked image modelling(MIM) with contrastive objectives \cite{zhang2024ctxmimcontextenhancedmaskedimage}. Although they have promising performance on several remote sensing tasks including change detection, landcover classification, and object detection, they neglect the temporal dimension of remote sensing data.

\paragraph{Domain Specifc Features}
While generic SSL methods often focus on raw pixel reconstruction or learning invariance to different augmentations, agricultural applications benefit from incorporating domain knowledge. For example, FGMAE \cite{wang2023featureguidedmaskedautoencoder} has shown that reconstructing agricultural indices(e.g:- NDVI) can improve downstream performance in MAE, compared to pixel-based reconstruction. Instead of focusing on reconstructing spectral indices, which primarily focus on spatial and spectral features, we focus on integrating both change related temporal features along with rich semantic spatial and spectral features, which provide rich spatial, temporal and spectral signal for agricultural monitoring.

\paragraph{Temporal Understanding in Remote Sensing}
Agricultural monitoring inherently requires rich understanding of temporal patterns. Recent works including U-Barn\cite{ubarn} and ALISE\cite{dumeur2024pavingwayfoundationmodels} emphasize the importance of effectively encoding satellite image time series to learn temporally-rich representations. In contrast, our work focuses on capturing temporally-informative features from a single acquisition at downstream, reducing reliance on dense temporal sequences.

\paragraph{Future Frame Prediction in other modalities}
Future frame prediction extends sequence modeling in time series data, aiming to generate future observations based on historical inputs. This task has been extensively studied across domains such as video understanding \cite{jang2024visualrepresentationlearningstochastic} and weather forecasting \cite{bodnar2024foundationmodelearth}. Recently, \cite{Ravirathinam2024ACI} explored a related task—predicting future frames from past visual sequences conditioned on weather data. In contrast, our work proposes a novel approach to future frame prediction in agricultural landscapes by leveraging only a single past frame, without relying on any external data. We hypothesize that this is feasible due to the inherently cyclical nature of agricultural patterns.

To the best of our knowledge, this is the first work to introduce temporal difference prediction and temporal frequency prediction as self-supervised pretext tasks for remote sensing imagery. Furthermore, we are the first to explore future frame prediction from a single past frame without incorporating any external information—an approach made viable by the cyclical patterns inherent in agricultural landscapes.

\section{Methodology}
\label{sec:methodology}
\begin{figure*}[t]
  \centering
   \begin{subfigure}{0.45\textwidth}
      \includegraphics[width=\textwidth]{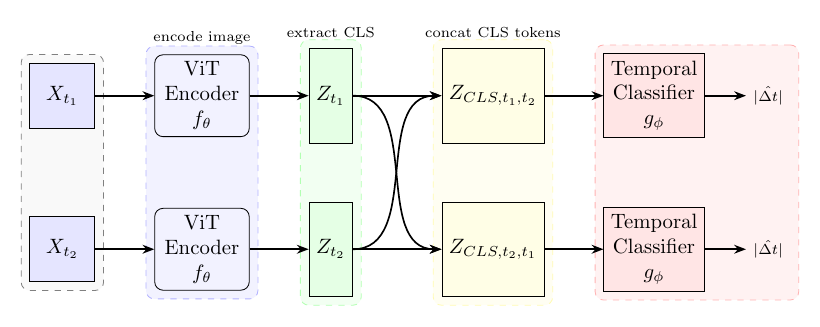}
      \caption{Time Difference Prediction (TD)}
      \label{fig:td}
    \end{subfigure}
    \begin{subfigure}{0.45\textwidth}
      \includegraphics[width=\textwidth]{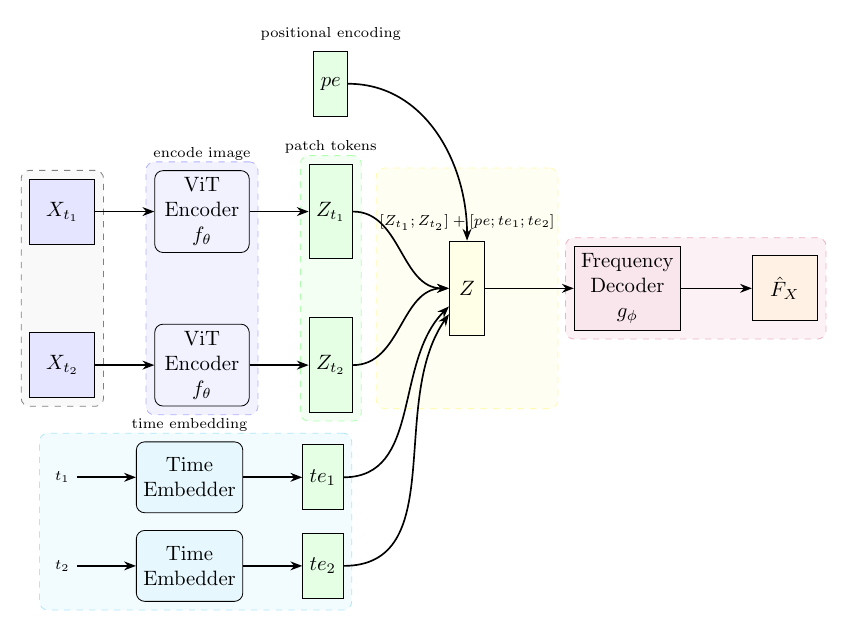}
      \caption{Temporal Frequency Prediction (FP)}
      \label{fig:fp}
    \end{subfigure}
    \begin{subfigure}{0.6\textwidth}
      \includegraphics[width=\textwidth]{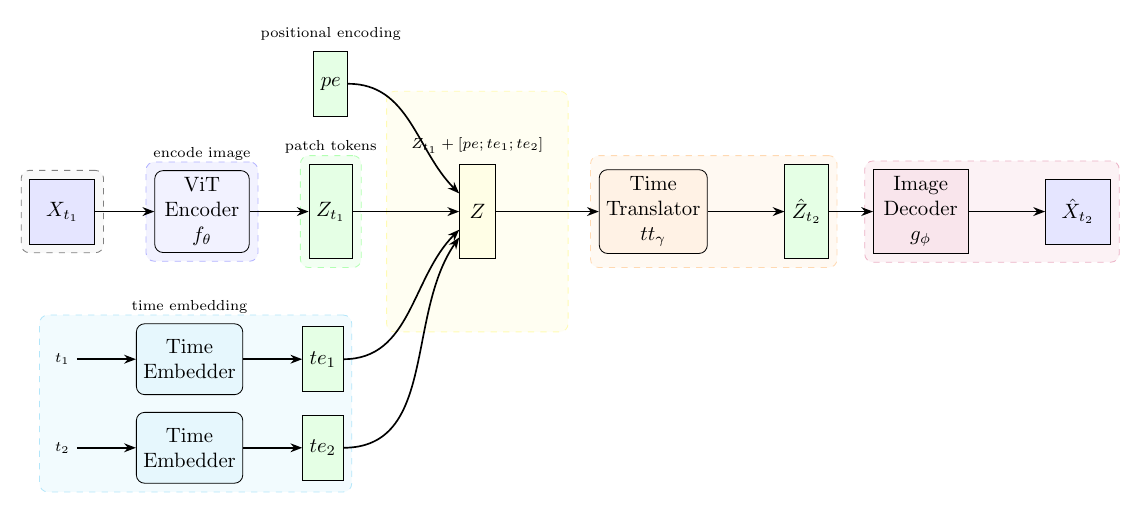}
      \caption{Future Frame Prediction (FF)}
      \label{fig:ff}
    \end{subfigure}
\caption{Overview of proposed pretext tasks. 
(a) \textbf{Time Difference Prediction (TD)}: Predicts absolute time gap $|\Delta t|$ from stacked CLS tokens $[(Z_{t_1})_{\text{CLS}}; (Z_{t_2})_{\text{CLS}}]$ via temporal classifier $g_\phi$. 
(b) \textbf{Temporal Frequency Prediction (FP)}: Predicts per-pixel dominant frequency map $\hat{F}_X$ from concatenated latents $[Z_{t_1}; Z_{t_2}] + [pe; te_1; te_2]$ using decoder $g_\phi$. 
(c) \textbf{Future Frame Prediction (FF)}: Predicts future frame $\hat{X}_{t_2}$ by generating latent $\hat{Z}_{t_2}$ from $Z_{t_1}$ and $[pe; te_1; te_2]$ using time translator $tt_\gamma$, followed by decoder $g_\phi$.
Here, $f_\theta$ is a ViT encoder, $g_\phi$ is a transformer decoder, and $tt_\gamma$ is a time translator module. $pe$ and $te_i$ denote positional and temporal encodings respectively.}
  \label{fig:methods_diagrams}
\end{figure*}

\subsection{Pretraining Data}
Our pretext tasks, as will be introduced later, require a dense time series of satellite imagery(SITS) over a target agricultural region. Existing large SITS pretraining datasets including SeCo\cite{Manas_2021_ICCV} and SSL4EO\cite{wang2023ssl4eos12largescalemultimodalmultitemporal} lack a definitive agricultural focus, and have sparse temporal sampling.
To effectively devise a pipeline to learn rich agricultural representations at both a regional scale and a national scale, we created two new SITS pretraining datasets with large temporal coverage, one at a regional scale of Tamil Nadu and another at the national scale of India. We used the already existing SICKLE and Fields of the World(FTW) benchmarks as our heuristic for sampling our training chips over the respective regions.

\subsubsection{Regional Pretraining}
\paragraph{SICKLE} \label{par:sickle_pretrain} SICKLE dataset provides multi-sensor imagery (Sentinel-2, Sentinel-1, and Landsat-8) from January 2018 till March 2021 over Cauvery Delta region in Tamil Nadu. They provide rich annotations for key agricultural tasks including crop type mapping, crop yield estimation, sowing date, transplanting date and harvesting date prediction. This dataset's agricultural focus and geographical specificity not only establish it as our primary downstream evaluation benchmark but also provide the empirical justification for concentrating our regional pretraining efforts on Tamil Nadu

\paragraph{Data Construction} To construct our regional scale pretraining dataset, we systematically identified all Sentinel-2 tiles that intersect with SICKLE's spatial extent, specifically tiles 44PKS, 44PKT, 44PLS, and 44PLT, with 44PLS and 44PLT having significant ocean coverage. After application of a water body mask, all 224x224 spatial chips were sampled and a dense SITS for each chip was constructed with a monthly resolution from the dataset's original acquisition period.

\paragraph{Dominant Frequency Map} \label{par:freq_constr} We additionally augument our regional dataset with a per-pixel dominant frequency map derived from the per-pixel NDVI time series in our data. Each per-pixel NDVI time series is interpolated to a regular monthly grid, and is smoothened using Savitzky-Golay filter. Dominant per-pixel fourier frequencies are calculated and stored as pixel values for each pixel sorted by their power. As discussed in \ref{sec:intro}, these frequencies capture different crop growth rates for crops grown in an agricultural parcel, and can naturally identify different parcels. We refer the reader to pseudocode~\ref{algo:dom_freq} for more details.
\begin{listing}[]%
\caption{Algorithm for Frequency Map Construction}
\begin{lstlisting}[language=python]
def construct_frequency_map(ndvi, timestamps, K=3):
    T, H, W = ndvi.shape
    # Calculate months since start
    months = [(t.year - timestamps[0].year) * 12 + t.month - timestamps[0].month for t in timestamps]
    reg_months = np.arange(min(months), max(months) + 1)

    # Interpolate Monthly NDVI
    ndvi_flat = ndvi.reshape(T, -1)
    reg_ndvi = np.zeros((len(reg_months), H * W))
    for i in range(H * W):
        f = interp1d(months, ndvi_flat[:, i], kind='linear', fill_value="extrapolate")
        reg_ndvi[:, i] = f(reg_months)

    # Smooth using Savitzky-Golay filter
    smoothed = savgol_filter(reg_ndvi, window_length=7, polyorder=4, axis=0)

    # FFT and frequency extraction
    fft_vals = fft(smoothed, axis=0)
    freqs = fftfreq(len(reg_months), d=1)
    amps = np.abs(fft_vals)

    # Select top-K positive frequencies
    pos = freqs > 0
    top_k = np.zeros((K, H * W))
    for i in range(H * W):
        idx = np.argsort(amps[pos, i])[-K:]
        top_k[:, i] = freqs[pos][idx]

    return top_k.reshape(K, H, W)
\end{lstlisting}
\label{algo:dom_freq}
\end{listing}%

\subsubsection{National Pretraining}
\paragraph{Fields of the World(FTW)} \label{par:ftw_pretrain} FTW is a field instance segmentation benchmark across multiple countries including India, and has over 70,000 chips globally and 1,960 chips specifically from India. Each chip consists of a bitemporal harmonized Sentinel 2 RGB+NIR imagery paired with instance and semantic segmentation masks, making it ideal to evaluate effectiveness of our methods at a national scale.

\paragraph{National Pretraining Data} To construct our national scale pretraining datasets, we referred to India chips from FTW(FTW India) and for each chip, we construct a 3x3 grid with chip at center and additionally sample its eight spatial neighbors, since an agricultural area is likely to be surrounded by other agricultural areas. Similar to our regional pipeline, we keep a monthly resolution but chose acquisition period from January 2016 to January 2019.

\subsection{Pretext Tasks}

\paragraph{Setup} \label{par:setup} We define our pretraining dataset as $S = (X_{t_{i}},X_{t_{j}},F_{X})$ consisting of bitemporal image pairs
, where $X_{t_{i}}$ and $X_{t_{j}}$ are satellite images at $t_{i}$ and $t_{j}$ with $t_{i} <= t_{j}$, and $F_{X}$ is per-pixel dominant frequency map constructed from entire time series. We define $f_{\theta}$ as a ViT encoder which extracts $Z_{t_{i}} \in \mathbb{R}^{(N+1) \times D^{in}}$ from $X_{t_{i}} \in \mathbb{R}^{H \times W \times C}$,where $N$ is the number of patches. From \citet{Vaswani2017AttentionIA}, we define positional encoding $PE(i,.)$
\begin{equation}
  \label{eq:posemb}
  PE(i,2j) = \sin(\frac{i}{B^{\frac{2j}{D}}})
\end{equation}
\begin{equation}
  PE(i,2j+1) = \cos(\frac{i}{B^{\frac{2j}{D}}})
\end{equation} 
where $i,j$ denote position and feature index, $D$ denotes length of the encoding and $B$ is a constant.
Similarly, we define time encoding of timestamp $t_{i}$ as $te_{i}$ as
\begin{equation}
  te_{i} = [PE(m_{i});PE(y_{i})]
\end{equation}
where $m_{i},y_{i}$ denote month, and year of $t_{i}$, $[;]$ denotes concatenation. We avoid using day-of-year\cite{Cong2022SatMAEPT} or month-of-year\cite{Tseng2023LightweightPT} encodings, as we believe they undermine inter-year differences.

\paragraph{Time Difference Prediction - TD} In this task (see Figure~\ref{fig:td}), we predict the absolute monthly time difference $|\Delta t| =|t_{2} - t_{1}|$ from latent representations $Z_{t_{1}}$ and $Z_{t_{2}}$ by applying a lightweight temporal classifier (e.g:-a three-layer MLP) on top of stacked CLS tokens from both representations
\begin{equation}
  Z_{CLS,t_{1},t_{2}} = [(Z_{t_{1}})_{CLS};(Z_{t_{2}})_{CLS}]
\end{equation}
where $(Z_{t_{i}})_{CLS}$ denote the CLS token of $Z_{t_{i}}$.
The classifer is trained to predict $|\Delta t|$ discretized into $C$ classes, and trained using a cross-entropy loss. 

To ensure robustness to temporal order, we use a symmetric loss function considering stacking $Z_{CLS,t_{1},t_{2}}$ and $Z_{CLS,t_{2},t_{1}}$ as input to the classifier. 

TD exploits temporal dimension of agricultural landscape by encoding temporal difference between sowing and harvest, or between two harvests in its representation. This is useful for learning crop phenological stages and growth patterns, which is important for agricultural monitoring.

\paragraph{Temporal Frequency Prediction - FP} FP (see Figure~\ref{fig:fp}) effectively addresses the limitations of TD, namely lack of dense supervision and more rich temporal features. By  predicting a per-pixel temporal frequency map $F_{X}$ from the latent representations $Z_{t_{1}}$ and $Z_{t_{2}}$, it allows encoder $f_{\theta}$ to learn dominant temporal features over each pixel from the entire time series.
We utilize MAE inspired transformer decoder $g_{\phi}$ to predict frequency map $\hat{F}_{X}$, by using concatenated latents(without CLS tokens) with summation of concatenated temporal encodings $te_{1}$ and $te_{2}$, and positional encoding $pe$.
\begin{equation}
  Z = [Z_{t_{1}};Z_{t_{2}}] + [pe;te_{1};te_{2}]
\end{equation}
\begin{equation}
  \hat{F}_{X} = g_{\phi}(Z)
\end{equation}    
This results in a rich temporal representation capturing temporal patterns(e.g:- crop rotation) beyond the immediate observable window. This is a regression task, and is optimized using normalized pixel-wise MSE\cite{He2021MaskedAA}.

\paragraph{Future Frame Prediction - FF} FF (see Figure~\ref{fig:ff}) is a generative task, where we predict the future frame $X_{t_{2}}$ from the past frame $X_{t_{1}}$. This addresses the limitation of dense temporal supervision, without the need of pre-computed frequency maps over the entire time series.
We retain the same transformer decoder $g_{\phi}$ as in FP, but additionally use another transformer decoder $tt_{\gamma}$, called time translator, to predict future latent representation $\hat{Z}_{t_{2}}$ from summation of $Z_{t_{1}}$ and concatenation of $te_{1}$, $te_{2}$ and $pe$. 
\begin{equation}
  \hat{Z}_{t_{2}} = tt_{\gamma}([Z_{t_{1}}] + [pe;te_{1};te_{2}])
\end{equation}
\begin{equation}
  \hat{X}_{t_{2}} = g_{\phi}(\hat{Z}_{t_{2}})
\end{equation}
Without the computational overhead of computing frequency maps, this results in rich temporal representation encoding casual relationship between current state and future state of an agricultural landscape. The network is similarly optimized using normalized pixel-wise MSE\cite{He2021MaskedAA} loss between predicted future frame $\hat{X}_{t_{2}}$ and ground truth $X_{t_{2}}$.

\section{Experiments \& Results}
\begin{table*}[t]
  \centering
    \caption{Performance comparison of our pretext tasks with MAE, pretrained DoFA, supervised ViT-S, and UNet3D on the SICKLE dataset. $\downarrow$ indicates lower is better, $\uparrow$ indicates higher is better. Bold values indicate the best performance for each task.}
  \label{tab:regional_eval}
  \setlength{\tabcolsep}{2mm}
  \begin{tabular}{lcccccc}
    \toprule
    \textbf{Model} & \textbf{Crop Type} & \textbf{Crop Yield} & \textbf{Yield In-Season} & \textbf{Sowing} & \textbf{Transpl.} & \textbf{Harvest} \\
     & (IoU $\uparrow$) & (MAPE $\downarrow$) & (MAPE $\downarrow$) & (MAPE $\downarrow$) & (MAPE $\downarrow$) & (MAPE $\downarrow$) \\
    \midrule
  FF            & \textbf{0.69595} & 0.36765          & 0.32516          & \textbf{0.015821} & 0.028394         & 0.068059 \\
  FP            & 0.61976          & \textbf{0.34338} & \textbf{0.30727} & 0.016081        & 0.022514         & 0.066166 \\
  TD            & 0.66628          & 0.36875          & 0.32877          & 0.017842          & 0.024985         & 0.065945 \\
  MAE           & 0.67138          & 0.37833          & 0.32701          & 0.017487          & 0.023406         & \textbf{0.065195} \\
    \midrule
  DoFA          & 0.58363          & 0.36144          & 0.34571          & 0.016849          & \textbf{0.022475}  & 0.071494 \\
    \midrule
  ViT-S         & \textbf{0.65523} & \textbf{0.36357} & 0.33687          & 0.016278          & 0.025533         & 0.069594 \\
  UNet3D        & 0.65442          & 0.38805          & 0.37964          & \textbf{0.012484} & \textbf{0.019249}& \textbf{0.047596} \\
    \bottomrule
  \end{tabular}
\end{table*}

\begin{table*}[t]
  \centering
    \caption{Performance comparison of different temporal gap between bitemporal pairs on FF. $\downarrow$ indicates lower is better, $\uparrow$ indicates higher is better. Bold values indicate the best performance for each task. FF denotes 0–3 months apart; FF(3) denotes 3 months apart; FF(3,6) denotes 3 and 6 months apart; FF(3,6,9) denotes 3, 6, and 9 months apart.}
  \label{tab:sickle_res2}
  \setlength{\tabcolsep}{2mm}
  \begin{tabular}{lcccccc}
    \toprule
    \textbf{Model} & \textbf{Crop Type} & \textbf{Crop Yield} & \textbf{Yield In-Season} & \textbf{Sowing} & \textbf{Transpl.} & \textbf{Harvest} \\
     & (IoU $\uparrow$) & (MAPE $\downarrow$) & (MAPE $\downarrow$) & (MAPE $\downarrow$) & (MAPE $\downarrow$) & (MAPE $\downarrow$) \\
    \midrule
  FF            & \textbf{0.69595} & 0.36765          & \textbf{0.32516} & 0.015821          & 0.028394         & 0.068059 \\
  FF-(3)        & 0.58737          & 0.34233          & 0.34674          & 0.01515           & 0.027323         & 0.066392 \\
  FF-(3,6)      & 0.64414          & \textbf{0.33411} & 0.36291          & \textbf{0.015033} & 0.025032         & 0.057168 \\
  FF-(3,6,9)    & 0.59340          & 0.34979          & 0.34940          & 0.01544           & \textbf{0.021851}& \textbf{0.05391}  \\
    \bottomrule
  \end{tabular}
\end{table*}

\begin{table*}[t]
  \centering
  \setlength{\tabcolsep}{4.5pt}
  \caption{Performance comparison of FF with MAE, pretrained DoFA, and supervised ViT-S on the India subset of the FTW dataset. FTW-India(all FTW countries except India), Ai4boundaries, France, and India refer to supervised pretraining regions, while metrics are reported on the India test set from FTW. Bold indicates the best performance per row.}
  \label{tab:national_results}
  \begin{tabular}{llccc}
    \toprule
    \textbf{Pretrain Region} & \textbf{Model} & \textbf{IoU ($\uparrow$)} & \textbf{Pixel Recall ($\uparrow$)} & \textbf{Object Recall ($\uparrow$)} \\
    \midrule
    \multirow{4}{*}{FTW - India}
  & FF & \textbf{0.541998} & \textbf{0.637823} & \textbf{0.029626} \\
  & MAE & 0.527783 & 0.628631 & 0.022341 \\
  & ViT-S & 0.493208 & 0.577795 & 0.022827 \\
  & DoFA & 0.38059 & 0.442879 & 0.002428 \\
    \midrule
    \multirow{4}{*}{Ai4boundaries}
  & FF & \textbf{0.530193} & \textbf{0.626151} & \textbf{0.023798} \\
  & MAE & 0.517006 & 0.615881 & 0.019427 \\
  & ViT-S & 0.415647 & 0.488697 & 0.0034 \\
  & DoFA & 0.368869 & 0.427868 & 0.001943 \\
    \midrule
    \multirow{4}{*}{France}
  & FF & \textbf{0.490628} & \textbf{0.577658} & \textbf{0.015542} \\
  & MAE & 0.465626 & 0.55264 & 0.014085 \\
  & ViT-S & 0.476051 & 0.573084 & 0.003885 \\
  & DoFA & 0.37447 & 0.439183 & 0.003885 \\
    \midrule
    \multirow{4}{*}{India}
  & FF & 0.492768 & 0.591907 & \textbf{0.024284} \\
  & MAE & \textbf{0.497104} & \textbf{0.60582} & 0.009228 \\
  & ViT-S & 0 & 0 & 0 \\
  & DoFA & 0.430635 & 0.513778 & 0.004371 \\
    \bottomrule
  \end{tabular}
\end{table*}

\begin{table*}[t]
  \centering
      \caption{Comparison of India FF with regionally pretrained TD, FP, FF, MAE; foundation model DoFA; and supervised ViT-S and UNet3D on the SICKLE dataset. $\downarrow$ indicates lower is better, $\uparrow$ indicates higher. Bold indicates the best score for each task.}
  \label{tab:scale_comp}
  \setlength{\tabcolsep}{2mm}
  \begin{tabular}{lcccccc}
    \toprule
    \textbf{Model} & \textbf{Crop Type} & \textbf{Crop Yield} & \textbf{Yield In-Season} & \textbf{Sowing} & \textbf{Transpl.} & \textbf{Harvest} \\
     & (IoU $\uparrow$) & (MAPE $\downarrow$) & (MAPE $\downarrow$) & (MAPE $\downarrow$) & (MAPE $\downarrow$) & (MAPE $\downarrow$) \\
    \midrule
  FF            & \textbf{0.69595} & 0.36765          & 0.32516          & 0.015821          & 0.028394         & 0.068059 \\
  FP            & 0.61976          & \textbf{0.34338} & \textbf{0.30727} & \textbf{0.016081} & 0.022514         & 0.066166 \\
  TD            & 0.66628          & 0.36875          & 0.32877          & 0.017842          & 0.024985         & 0.065945 \\
  MAE           & 0.67138          & 0.37833          & 0.32701          & 0.017487          & 0.023406         & 0.065195 \\
    \midrule
  \midrule
  India FF      & 0.60483          & 0.39750          & 0.37407          & 0.016549          & \textbf{0.022403} & \textbf{0.064496} \\
    \midrule
  DoFA          & 0.58363          & 0.36144          & 0.34571          & 0.016849          & 0.022475         & 0.071494 \\
    \midrule
  ViT-S         & \textbf{0.65523} & 0.36357          & 0.33687          & 0.016278          & 0.025533         & 0.069594 \\
  UNet3D        & 0.65442          & \textbf{0.38805} & \textbf{0.37964} & \textbf{0.012484} & \textbf{0.019249}& \textbf{0.047596} \\
    \bottomrule
  \end{tabular}
\end{table*}

We aim answer two fundamental questions that emerge from our core hypothesis that agricultural landscape's cyclical nature can inform better self-supervised pretext tasks. Specifically, we aim to answer: do temporal pretext tasks learn more agriculturally relevant representations, and which temporal characteristicis are most valuable for different agricultural monitoring tasks. Additionally, we also seek to understand the role of national scale pretraining on tackling regional challenges.

\subsection{Downstream Evaluations}
\paragraph{SICKLE} SICKLE benchmark, as discussed earlier in Section~\ref{par:sickle_pretrain}, is a comprehensive agricultural monitoring benchmark over Cauvery Delta region of Tamil Nadu, consisting of SITS with annotations for crop type mapping, crop yield estimation, and sowing, transplanting and harvesting date prediction. Crop type mapping is a binary segmentation task for segmenting paddy pixels, measured by IoU, while crop yield estimation estimates the yield of a paddy parcel, measured by mean-absolute-percentage-error(MAPE). Similar to \citet{Sani_2024_WACV}, we consider using both using satellite images during actual growing season, referred to as in-season, as well as using the whole time series available for estimating crop-yield. Date prediction tasks are carried out using in-season time series, where the task is to predict how many days apart from reference does sowing, transplanting and harvesting occur. Date prediction, similar to yield estimation, is a regression tasks and evaluated using MAPE. 

\paragraph{Fields of the World} FTW, as discussed earlier in Section~\ref{par:ftw_pretrain}, consists of over 70,000 chips with annotations for instance and semantic segmentation mask. Similar to \citet{kerner2024fieldsworldmachinelearning}, we consider field instance segmentation performance on India for 3-class setting: field interior, field exterior and field boundary. Our downstream task use IoU, Pixel Recall and Object Recall as our metrics due to availability of presence-only labels in FTW India. We consider evaluation strategy similar to \citet{kerner2024fieldsworldmachinelearning}, which involved three settings: training on all FTW countries following by finetuning only on India, training on Ai4boundaries countries followed by finetuning on India, and training on France followed by finetuning on India. In addition, we also consider training on India only and evaluation on India test set of FTW.

\subsection{Implementation Details}
We utilize a ViT-S(small) encoder for our pretraining experiments, and use a 3-layer MLP classifier for TD, 6-layer Tranformer decoder for FP, and 2-layer Time Translator followed by 4-layer Decoder for FF. For FP, we consider the top-3 dominant temporal frequencies. Unless otherwise specified, bitemporal pairs are sampled within a maximum temporal gap of three months. Additionally, we consider an MAE baseline trained on single frame reconstruction on our pretraining data. All our pretraining and downstream experiments use 4-band(red,blue,green and nir) imagery, in alignment with the FTW dataset, which provides only these four bands. We consider DoFA as our pretrained RSFM baseline, due to its native support for our 4-band imagery, and its competitive or superior performance to most RSFMs across a range of remote sensing downstream tasks.
For the SICKLE downstream task, we use a UPerNet segmentation head and compare against both supervised ViT-S and supervised UNet3D baselines. For the FTW India downstream task, we employ a simpler segementation head constituting three ConvTranpose-Conv-BN-Relu-Dropout layers, followed by a final ConvTranpose-Conv layer, using ConvTranspose for feature upsampling, and Conv for feature refinement. For FTW India, we additionally compare against supervised ViT-S.
All our SSL experiments use AdamW with cosine annealling schedule, trained for 100 epochs.  We use a batch size of 256 for regional-scale pretraining, and an effective batch size of 640 with 10-epoch warmup for national-scale pretraining. All our SSL and downstream experiments are run on a single RTX A6000(48GB) GPU, except our national scale pretraining use 2 x A100(40GB) GPUs.

\subsection{Regional Scale Evaluation on SICKLE}

\paragraph{Do Temporal Pretext Tasks perform better?}As evident from Table~\ref{tab:regional_eval}, our temporal pretext tasks demonstrate clear advantages over existing methods. FF achieves the highest IoU of 69.6\% on crop type mapping, substantially outperforming DoFA (58.4\%), MAE (67.1\%) and supervised baselines(65.5\%). Similarly, FP achieves the lowest MAPE of 34\% for crop yield and 31\% for in-season prediction, outperforming all supervised and self-supervised baselines. While temporal tasks remain competitive with our baselines on date prediction tasks, they show consistent advantages across the broader spectrum of agricultural monitoring applications. This suggests that learning temporal characteristics of agricultural landscapes leads to more informative and transferable agricultural representations than spatial-only or globally-pretrained alternatives.

\paragraph{Which Temporal Pretext Task for Agriculture? } 
Among our temporal pretext tasks, FF and FP show complementary strengths aligned with different agricultural monitoring objectives. FF outperforms FP and TD by 7.6\% and 3.0\% respectively on crop type mapping, while FP achieves superior performance for crop yield prediction, outperforming both FF and TD by 2.4\% and 2.5\% MAPE points respectively on both in-season and complete season prediction. This performance differentiation reveals an important principle: FF learns comprehensive spatial, temporal and spectral crop characteristics suitable for identification tasks, while FP leverages its global temporal horizon to capture growth dynamics critical for yield estimation. TD shows intermediate performance across tasks, suggesting that simple temporal differences provide valuable but less specialized representations compared to generative (FF) or frequency-based (FP) approaches.

\paragraph{Why Different Downstream Favor Different Temporal Pretexts?}
The task-specific performance patterns reflect fundamental differences in different agricultural monitoring applications. While FF's comprehensive understanding of temporal evolution over shorter horizon allows its to capture and anticipate rich spatial, phenological, and spectral patterns suitable for crop mapping, it remains suboptimal for anticipating crop stresses. On the other hand FP's frequency analysis approach captures both seasonal growth patterns and stress events over a much longer horizon, allowing it to learn robust representations for detecting crop stresses and yield estimation. Although neither beats supervised baselines on date prediction tasks, suggesting supervised approaches still remain superior for precise event detection.

\paragraph{Does Time Scale Matter in Agriculture?} 
Although the previous result shed a light on this, we conducted an experiment on effect of different temporal gaps on performance of FF. Table~\ref{tab:sickle_res2} reveals a clear principle: different agricultural monitoring tasks align with specific temporal scales in agricultural systems. 
While crop type mapping achieves best performance (69.6\% IoU) with FF-(0,1,2,3), suggesting that crop identification benefits from observing diverse phenological stages within growing seasons—early growth, peak vegetation, and senescence phases,
crop yield prediction performs optimally (33.4\% MAPE) with FF-(3,6) configuration, indicating that yield estimation requires understanding seasonal to inter-seasonal patterns that capture complete growing cycles and stress period impacts on final productivity. Date prediction tasks shows progressive improvement with longer temporal contexts, achieving best performance (3.01\% MAPE) with FF-(3,6,9), suggesting that accurate agricultural timing prediction benefits from understanding multi-season planting patterns.

\subsection{National Scale Evaluation on FTW India}

\paragraph{Which Pretext Task to Scale?} Our regional scale experiments(see Table~\ref{tab:regional_eval}) suggests FF leads to better crop type representations, while FP performs the best on crop yield estimation. Our intuition that field boundary delineation is more closer to crop type mapping instead of yield estimation led us to go forward with FF for our national scale training.

\paragraph{Do Temporal Pretext Task improve Cross-Region Transfer?} Table~\ref{tab:national_results} demonstrates that FF consistently outperforms MAE across different geographic supervised pretraining sources, with advantages ranging from 1.3\%(Ai4boundaries) to 2.5\%(France), suggesting that FF learn more fundamental agricultural principles that transcend specific geographic contexts.

\paragraph{Does SSL eliminate the need for Cross Region data?}  
FF and MAE on India supervised pretraining only, recovers 99.7\% and 100.8\% of the supervised performance of FTW-India suggesting SSL pretraining alleviate the need of cross-region labelled data in label sparse setting.

\subsection{Role of Geographical Scale}
Foundation model paradigms suggest that larger, more diverse datasets produce superior representations. However, agricultural systems are inherently location-specific due to climate, soil, crop varieties, and management practices. To answer this, question, we evaluated our national scale FF pretrained representations(India FF) on the SICKLE benchmark.

\paragraph{Regional representations substantially outperform National representations}
Table~\ref{tab:scale_comp} reveals a striking result that challenges foundation model conventions. Regional pretraining achieves 9.1\% advantage for crop type mapping, 3.0\% advantage for crop yield prediction, and consistent improvements for phenological dating. This finding suggests that agricultural foundation models should prioritize geographic specialization over scale, contrary to conventional computer vision wisdom. The location-specific nature of agricultural systems—climate, crops, and practices—creates domain characteristics that benefit more from regional focus than global generalization.

\section{Discussion}
Our work demonstrates that incorporating domain knowledge into pretext task design can substantially improve representation learning. This challenges the prevailing approach of adapting generic computer vision methods to specialized domains. The discovery that regional pretraining outperforms national scale challenges foundation model orthodoxy. Agricultural landscapes provide an ideal testbed for temporal representation learning because their dynamics are both complex(multi-scale) and predictable (seasonal, cyclical).

\section{Conclusion}
This work demonstrates that agricultural landscapes's inherent temporal structure provides powerful supervisory signals for self-supervised learning. By designing three novel pretext tasks that leverages this structure, we achieve substantial improvements over diverse agricultural monitoring tasks. Further, this work shows that regional pretraining substantially outperforms national scale for regional downstream tasks, emphasizing on development of regional agricultural representations instead of global representations for tackling regional agricultural challenges.

\section*{Acknowledgments} 
We thank the faculty and research staff at Plaksha University for their insightful feedback during various stages of this work. We are grateful to the team behind the SICKLE and Fields of the World datasets for publicly releasing high-quality annotated data, which was instrumental in our experiments. This research was supported by the SRG019 startup grant from Plaksha University.
\bibliography{refs}

\appendix

\section{Regional Pretraining Data Construction}
To construct our regional-scale pretraining dataset for Tamil Nadu, India, we utilized the SICKLE dataset \cite{Sani_2024_WACV}, a comprehensive benchmark for agricultural monitoring in the Cauvery Delta region of Tamil Nadu. The SICKLE dataset provides multi-sensor imagery from Sentinel-1, Sentinel-2, and Landsat-8 satellites, along with detailed annotations for key agricultural parameters such as crop type, phenological dates (sowing, transplanting, harvesting), and crop yield at multiple spatial resolutions (3m, 10m, and 30m). The dataset spans the time period from January 2018 to March 2021, making it ideal for capturing the temporal dynamics of agriculture in the region.

To construct the pretraining dataset, we used the spatial and temporal extent of the SICKLE dataset to sample Sentinel-2 L2A imagery. The following steps outline our methodology:

\begin{figure}[H]
  \centering
  \includegraphics[width=0.4\textwidth]{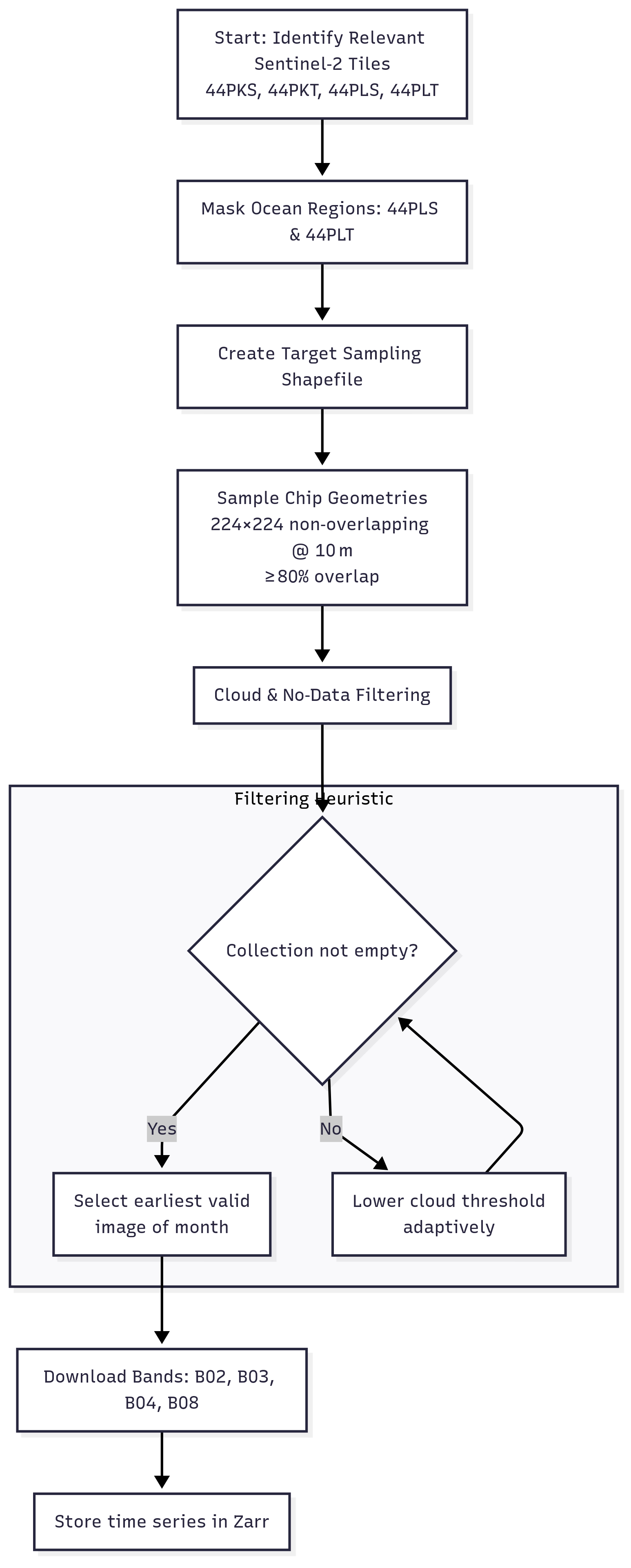}
  \caption{This figure illustrates the complete workflow of regional pretraining data collection.}
  \label{fig:regional_ds}
\end{figure}

\begin{enumerate}
    \item \textbf{Identifying Relevant Sentinel-2 Tiles}: The spatial extent of SICKLE dataset overlaps four Sentinel-2 tiles: 44PKS, 44PKT, 44PLS, and 44PLT.
    \item \textbf{Masking Ocean Regions}: Among the identified tiles, \textbf{44PLS} and \textbf{44PLT} contain significant ocean coverage, which exhibits temporal dynamics vastly different from agricultural land. To ensure the dataset focuses on relevant agricultural areas, we masked out the ocean regions and created a shapefile representing the target sampling region.
    \item \textbf{Sampling Sentinel-2 Chips Geometries}: Using the target shapefile, we exhaustively sampled \textbf{224 x 224 non-overlapping chip geometries} at a spatial resolution of \textbf{10m}. To ensure the chips were representative of the agricultural areas, we retained only those with at least \textbf{80\% overlap} with the target shapefile.
    \item \textbf{Cloud and NoData Filtering}: For each sampled chip geometry, we downloaded Sentinel-2 L2A Scene Classification Layer(SCL) for time period spanning \textbf{January 1, 2018, to April 1, 2021}, which aligns with the temporal coverage of the SICKLE dataset. For each chip, we applied the following heuristic: \begin{itemize}
        \item We prioritized selecting the \textbf{earliest image in the month} that satisfied a predefined threshold for cloud and nodata.
        \item If no image within the month met the threshold, we lowered the threshold for that specific region.
        \item This adaptive filtering approach allowed us to identify relevant Sentinel-2 chips to balance data quality and temporal coverage, ensuring that each chip had at least one valid image per month.
    \end{itemize}
    \item \textbf{Data Acquisiton}: For the identified timestamps for each chip geometry, we acquired B02, B03, B04,and B08 bands and stored the time series in a Zarr store. The resulting dataset is high-resolution monthly time series tailored for temporal pretext tasks.
\end{enumerate}
\section{National Pretraining Data Construction}
For our national-scale experiments, we utilized India subset of the Fields of the World(FTW) dataset\cite{kerner2024fieldsworldmachinelearning}, which provides agricultural field instance segmentation annotations across diverse regions in India. FTW includes over 70,000 samples globally with 1,960 samples specifically from India(of which 399 samples are held-out for testing).Each sample in FTW includes bitemporal harmonized Sentinel-2 RGB+NIR imagery paired with instance and semantic segmentation masks 
, making it an ideal resource for agricultural monitoring at a national scale.

To construct our national-scale pretraining dataset, we extended the India subset of FTW by increasing its temporal resolution, as well as sampling additional neighboring regions around each chip geometry. The following steps outline our methodology:

\begin{figure}[H]
  \centering
  \includegraphics[width=0.5\textwidth]{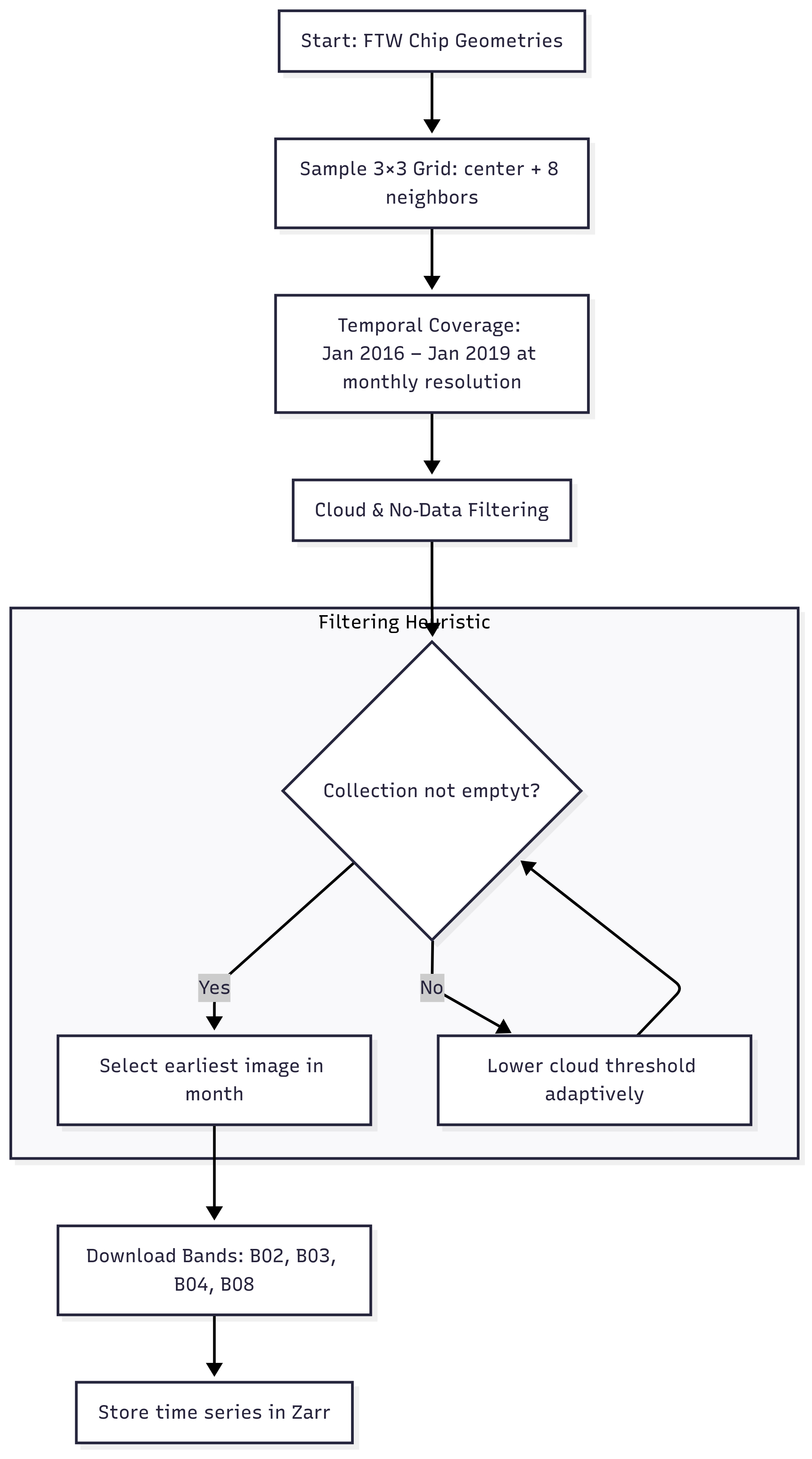}
  \caption{This figure illustrates the complete workflow of national pretraining data collection.}
  \label{fig:national_ds}
\end{figure}

\begin{figure*}
  \centering
  \includegraphics[width=0.95\textwidth]{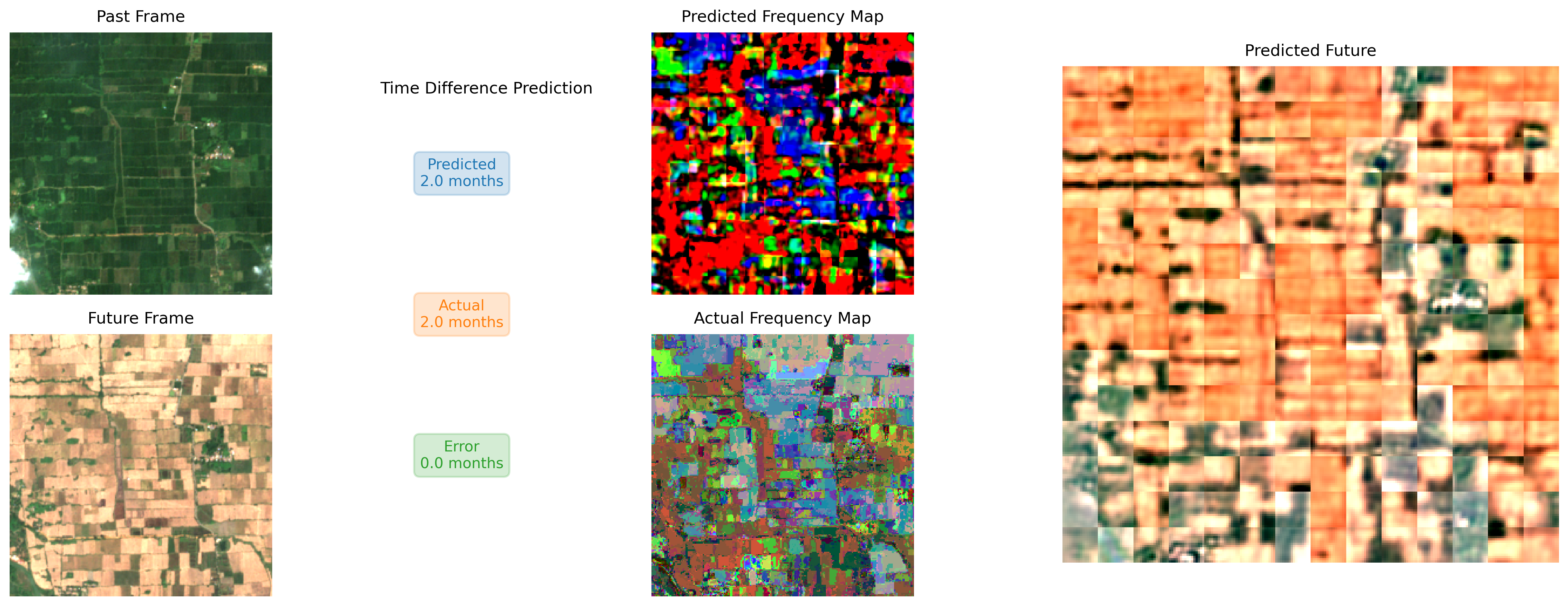}\par\vspace{2mm}
  \includegraphics[width=0.95\textwidth]{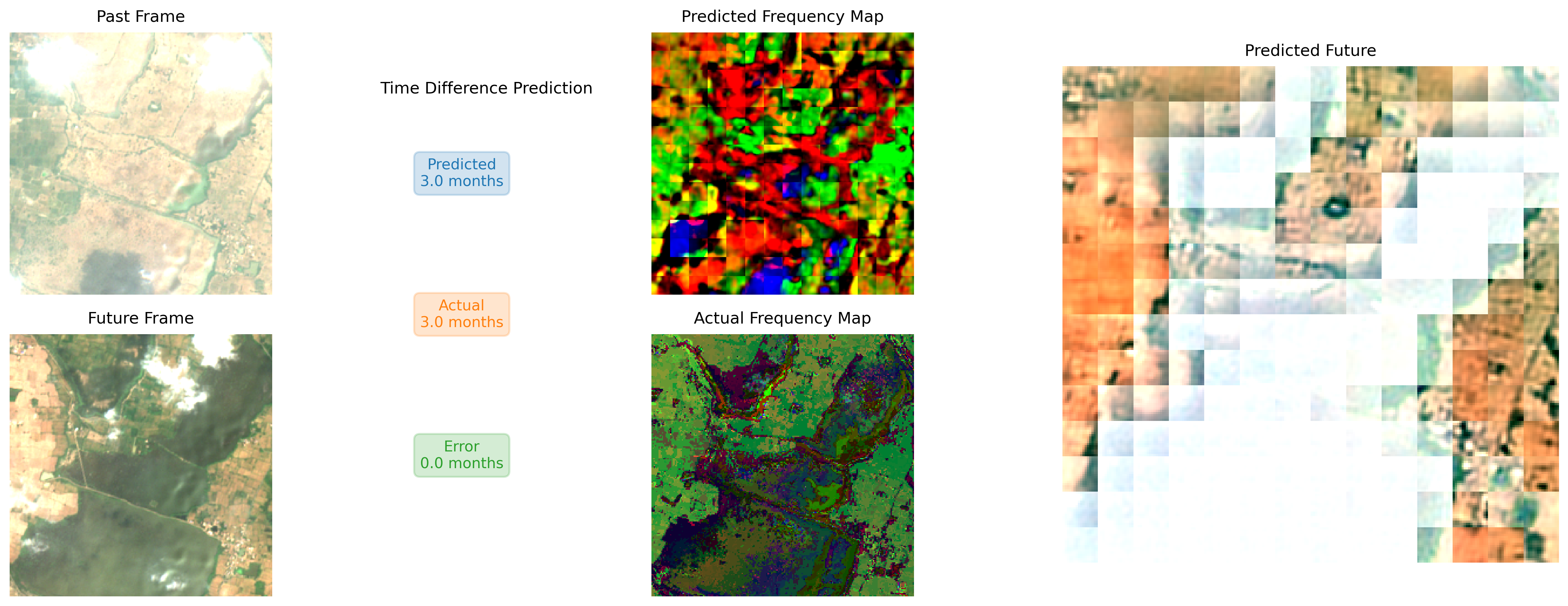}\par\vspace{2mm}
  \includegraphics[width=0.95\textwidth]{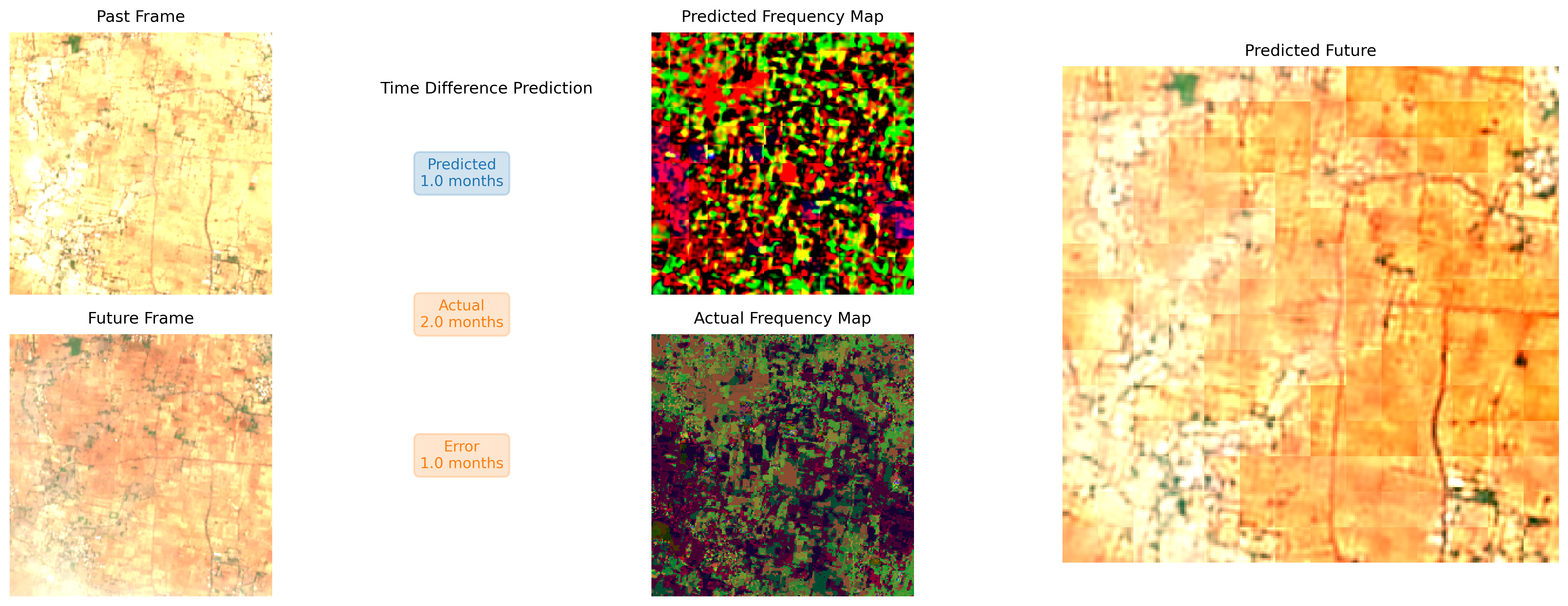}
  \caption{
    Visualization of predictions by TD, FP, and FF models across different samples.
    (a) The predicted future aligns well with the actual future; the frequency map naturally delineates farm parcels. 
    (b) The field appears to be undergoing flood irrigation, which the models fail to fully anticipate. 
    (c) The TD model's prediction is offset by one month, likely due to minimal visual change between frames.
  }
  \label{fig:stacked_images}
\end{figure*}

\begin{enumerate}
    \item \textbf{Sampling Neighboring Chips}: FTW provides predefined chip geometries, which cover diverse regions in India. For each chip in the dataset, we sampled its eight neighboring chips(top,bottom,left,right,top-right,top-left,bottom-right,bottom-left). This resulted in a \textbf{3x3 grid around each FTW sample}. The rationale behind this approach was that areas surrounding agricultural fields are likely to be agricultural fields.
    \item \textbf{Temporal Coverage}: Similar to regional-scale dataset, we kept monthly temporal resolution, but kept acqusition period from January 2016 to January 2019.
    \item \textbf{Cloud and No-Data Filtering}: To ensure high-quality imagery, we used the same filtering heuristic: \begin{itemize}
        \item For each chip in 3x3 grid, we prioritized selecting \textbf{the earliest image in the month} that satisfied a predefined threshold for valid pixels
        \item If no image within the month met the threshold, we \textbf{lowered the threshold adaptively} for that specific region to ensure temporal continuity while maintaining data quality.
    \end{itemize}
    \item \textbf{Data Acquisiton}: For each filtered chip in the 3x3 grid, we downloaded B02, B03, B04,and B08 bands and stored the time series in a Zarr store.
\end{enumerate}

\section{Visualization of Results}

Figure~\ref{fig:stacked_images} presents qualitative examples of our model's pretext predictions:

\begin{enumerate}
  \item[a.] \textbf{Well‑behaved growth cycle}: All three models accurately predict the crop’s progression, and the FP model’s frequency map (middle column) effectively delineates individual farm parcels even under homogeneous vegetation cover.
  \item[b.] \textbf{Flood irrigation pattern}: The ground truth (left) shows a flood irrigation event. FF partially capture these water pulses.
  \item[c.] \textbf{Subtle phenological changes}: In this image, the TD model’s forecast is shifted by one month—likely due to less visual change between the two time points.
\end{enumerate}
\end{document}